%% file: main.tex
\documentclass[letterpaper, 10 pt, conference]{ieeeconf}  % Comment this line out if you need a4paper

\IEEEoverridecommandlockouts                              % This command is only needed if 
\usepackage{mathrsfs}
\usepackage{amsfonts}
\usepackage{graphics} 
\usepackage{graphicx}
\usepackage[normalem]{ulem}
\usepackage{hyperref}
\usepackage{multirow}
\usepackage{xcolor}
\usepackage{pifont}
\usepackage{booktabs}
\usepackage{makecell}
\usepackage{colortbl}

\definecolor{impro}{RGB}{210,210,210}

\newcommand{\cc}{\color[rgb]{0,0.6,0.3}\ding{52}}
\newcommand{\xx}{\color[rgb]{0.6,0,0}{\ding{55}}}

\newcommand{\new}[1]{{\textcolor[rgb]{0,0,0}{#1}}}

\makeatletter
\newcommand{\printfnsymbol}[1]{%
        \textsuperscript{\@fnsymbol{#1}}%
}
\makeatother

\makeatletter
\DeclareRobustCommand\onedot{\futurelet\@let@token\@onedot}
\def\@onedot{\ifx\@let@token.\else.\null\fi\xspace}

\makeatother

\begin{document}

\title{\LARGE \bf
BlinkFlow: A Dataset to Push the Limits of Event-based Optical Flow Estimation
}

\author{Yijin Li$^{1}$\printfnsymbol{1}\thanks{\printfnsymbol{1}Yijin Li and Zhaoyang Huang contributed equally to this work.}, Zhaoyang Huang$^{2}$\printfnsymbol{1}, Shuo Chen$^{1}$, Xiaoyu Shi$^{2}$, Hongsheng Li$^{2}$, Hujun Bao$^{1}$, \\Zhaopeng Cui$^{1}$, Guofeng Zhang$^{1}$\printfnsymbol{2}\thanks{\printfnsymbol{2}Guofeng Zhang is the corresponding author.} \\ 
$^{1}$State Key Lab of CAD\&CG, Zhejiang University \\
$^{2}$Multimedia Laboratory, The Chinese University of Hong Kong 
}

\maketitle
\thispagestyle{empty}
\pagestyle{empty}

\begin{abstract}

Event cameras provide high temporal precision, low data rates, and high dynamic range visual perception, which are well-suited for optical flow estimation. While data-driven optical flow estimation has obtained great success in RGB cameras, its generalization performance is seriously hindered in event cameras mainly due to the limited and biased training data. In this paper, we present a novel simulator, BlinkSim, for the fast generation of large-scale data for event-based optical flow. 
BlinkSim incorporates a configurable rendering engine alongside an event simulation suite.
By leveraging the wealth of current 3D assets, the rendering engine enables us to automatically build up thousands of scenes with different objects, textures, and motion patterns and render very high-frequency images for realistic event data simulation. Based on BlinkSim, we construct a large training dataset and evaluation benchmark BlinkFlow that contains sufficient, diversiform, and challenging event data with optical flow ground truth. Experiments show that BlinkFlow improves the generalization performance of state-of-the-art methods by more than 40\% on average and up to 90\%. Moreover, we further propose an Event-based optical Flow transFormer (E-FlowFormer) architecture. Powered by our BlinkFlow, E-FlowFormer outperforms the SOTA methods by up to 91\% on the MVSEC dataset and 14\% on the DSEC dataset and presents the best generalization performance. The source code and data are available at \url{https://zju3dv.github.io/blinkflow/}.

\end{abstract}

\section{INTRODUCTION}

Event cameras detect changes in intensity at each pixel of the image as a stream of asynchronous events, which naturally depicts pixel motions.
Besides, their high temporal precision (on the order of microseconds), low data rates, and high dynamic ranges ($>$ 120 dB) far exceed traditional RGB cameras.
Such significant characteristics make event cameras popular in robotics~\cite{event_quadrotor,event_obstacle} and particularly suitable for tasks such as optical flow~\cite{eraft,secrets} estimation, which requires accurate and efficient tracking of pixel motion between frames.

\begin{figure}[!t]
\begin{center}
    \includegraphics[width=1.0\linewidth]{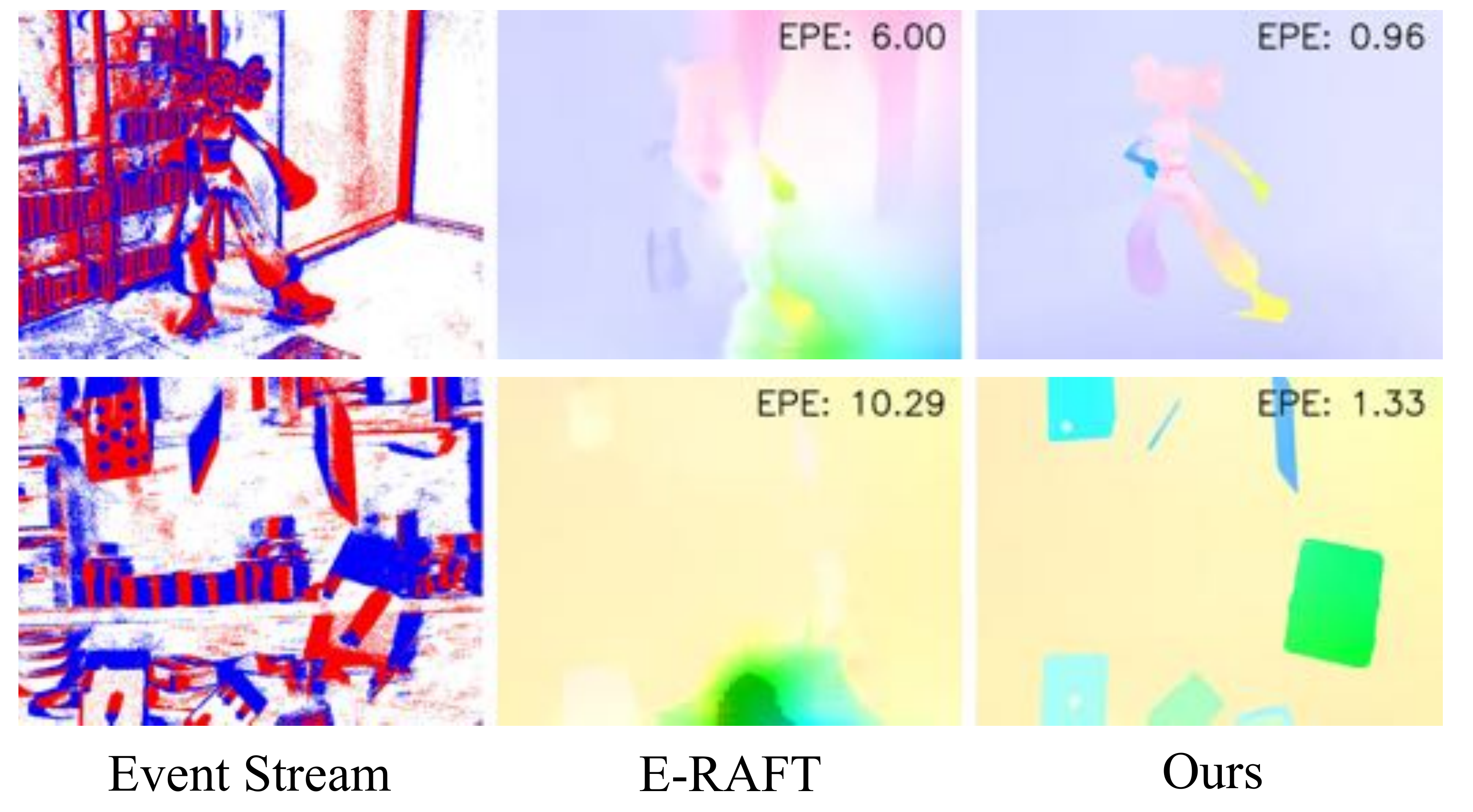}
\end{center}
\vspace{-1.0em}
\caption{Comparison of generalization. E-RAFT~\cite{eraft} overfits the DSEC dataset~\cite{dsec} and generalizes poorly to unseen environments. In contrast, our E-FlowFormer powered by the proposed BlinkFlow dataset presents good generalization performance and recovers complex and flexible optical flows.
}
\vspace{-1.0em}
\label{fig:teaser}
\end{figure}

\begin{figure}[!t]
\begin{center}
    \includegraphics[width=1.0\linewidth]{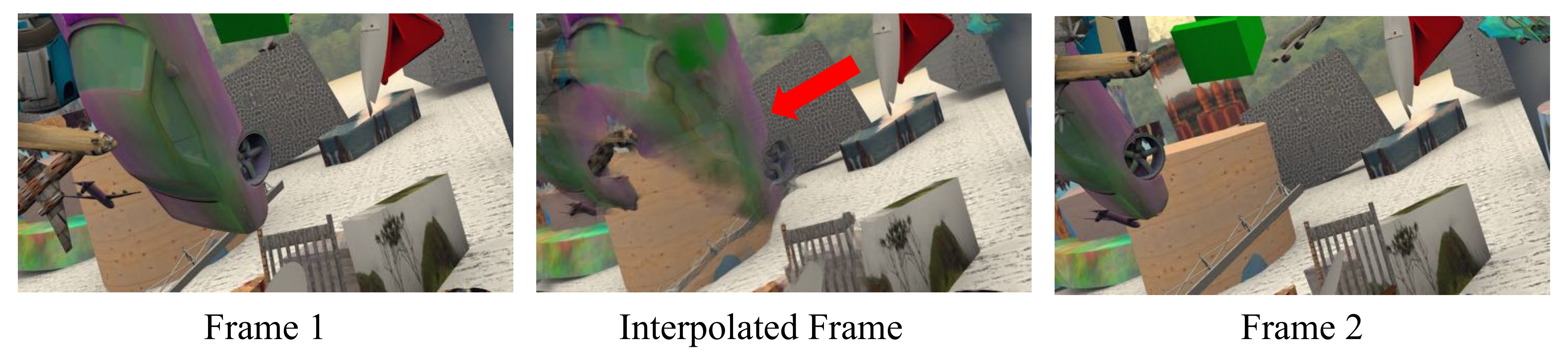}
\end{center}
\vspace{-1.0em}
\caption{Video interpolation artifacts (pointed by red arrows).
}
\vspace{-1.0em}
\label{fig:interpolation}
\end{figure}

The success of data-driven optical flow estimation for RGB cameras motivates researchers to develop learning-based methods for event-based optical flow estimation.
However, previous works~\cite{ev_flownet,spike_flownet,eraft} highly overfit their dataset (Fig.~\ref{fig:teaser}) and their performance is far from satisfactory in practical scenarios.
One major reason is the lack of large-scale diversiform training data along with a benchmark to evaluate their generalization performance.
Due to the novelty and complexity of the event camera, there are currently only four event camera datasets that contain the ground truth data of optical flow. These datasets cover limited scenarios, simple motion patterns and only contain sparse optical flow ground truth, as shown in Table~\ref{tab:dataset}.
To alleviate the problem, one potential strategy~\cite{video2event} is converting the existing RGB frame-based dataset into event data based on video interpolation, but it inevitably produces artifacts and introduces data bias.
Such issues may be acceptable in object recognition~\cite{event_recognition} and semantic segmentation~\cite{video2event}, which focus on high-level semantics, but can severely disturb low-level tasks such as optical flow especially when encountering violent motions and occlusion, as shown in Fig.~\ref{fig:interpolation}.

\begin{figure*}[!t]
\begin{center}
    \includegraphics[width=0.9\linewidth]{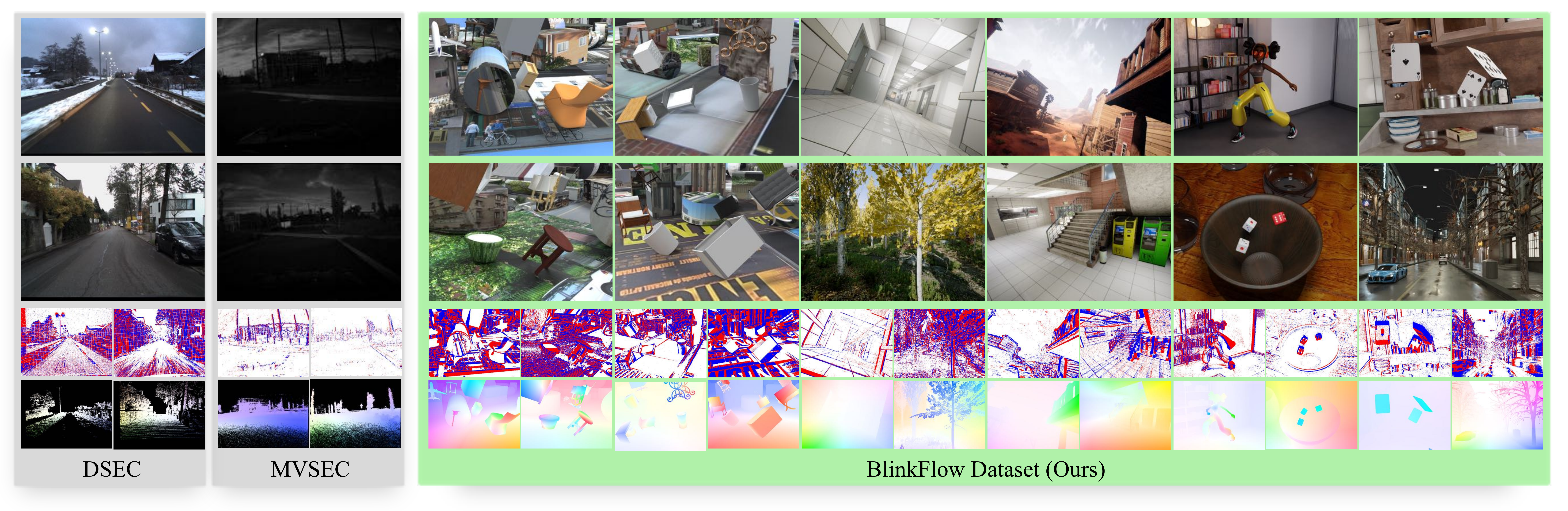}
\end{center}
\vspace{-1.0em}
\caption{Example scenes from DSEC, MVSEC and the proposed BlinkFlow dataset. \textbf{3rd row: Event data}, \textbf{4th row: Optical flow images}. Our BlinkFlow dataset contains complex object/camera motions and various scenarios which significantly outperform DSEC and MVSEC in quantity, quality, and diversity. The 1-2, 3-4, and 5-6 columns of the BlinkFlow Dataset correspond to the sequences of FlyingObjects, E-Tartan and E-Blender, respectively. Best viewed on a color screen in high resolution.}
\label{fig:dataset}
\end{figure*}

We observe that the favorable outcome of frame-based optical flow learning may largely be attributed to the synthetic datasets~\cite{flyingthings,flyingchair}.
Motivated by them, we propose BlinkSim, an event simulator that renders event data sequences and obtains their optical flow ground truth.
BlinkSim consists of two modules including a configurable rendering engine and a simulation suite with integrated multiple event emulators.
Our effort is similar in spirit to the FlyingChair~\cite{flyingchair} and FlyingThings~\cite{flyingthings} for RGB image-based optical flow estimation, but it is customized for event cameras.
Specifically, BlinkSim allows rendering very high frame rate images and is coupled with the event data simulator suite, which ensures realistic event data generation.
Based on BlinkSim, we generate a large training dataset and evaluation benchmark BlinkFlow.
As shown in Fig.~\ref{fig:dataset}, BlinkFlow covers diverse scene categories and complex motion patterns with sufficient realism.
To boost the research in event camera communities,  we also open-source the BlinkSim besides the release of BlinkFlow data. BlinkSim can be extended to alleviate the data problem in other tasks related to event cameras, such as feature tracking~\cite{eklt,bian2023context} and depth estimation~\cite{event_depth}.

Aside from data issues, the unfamiliar way in which events encode visual information also poses challenges for optical flow estimation.
Recent methods~\cite{eraft,ste_flownet} have demonstrated that the methods designed for images, i.e., usage of correlation volume, can be adopted and work quite well on event data.
However, as event cameras provide neither absolute brightness information nor spatially continuous data and the event data suffers from severe noise compared to image data, 
directly applying correlation to event data cannot produce a discriminative matching cost.
To address this problem, we propose an Event optical Flow transFormer (E-FlowFormer) architecture, which enhances the event feature extraction by fully encoding information from input events.
Empowered by the diversiform and high-quality training data of our BlinkFlow, E-FlowFormer learns effective and generalizable event features to build a distinct correlation volume for the following flow refinement.

Our contributions can be summarized in three folds: 1) We develop a novel simulator, BlinkSim, to synthesize events along with related ground truth. 2) We build a dataset BlinkFlow based on BlinkSim, for training and evaluating event-based optical flow estimation methods. BlinkFlow exceeds previous datasets in quantity, quality, and diversity.
3) We propose an E-FlowFormer to fully unleash the power of the large-scale training data provided by BlinkFlow. E-FlowFormer ranks 1st on mainstream benchmarks DSEC and MVSEC and the proposed BlinkFlow benchmark.

\input{tables/dataset}

\section{RELATED WORK}
\subsection{Optical Flow Dataset \new{and Simulator} for Event Camera}
As far as we know, there are currently only four public event camera datasets that include optical flow ground truth data. 
which can be categorized based on how the ground truth optical flow is obtained.
In DVSFLOW16~\cite{DVSFLOW16} and DVSMOTION20~\cite{event_surface},
\new{the ground truth optical flow was deduced from the camera's rotational movement, as measured by the inertial measurement unit.}
DSEC~\cite{dsec} and MVSEC~\cite{ev_flownet} computed the ground truth through LiDAR SLAM.
Both ways limit the ground truth to the static elements of the scene and the latter one only offers sparse data up to a specific distance and height.
Besides, these datasets have a very limited motion pattern. For example, DVSFLOW16~\cite{DVSFLOW16} and DVSMOTION20~\cite{event_surface} contain only rotational camera motion,
\new{while in DSEC~\cite{dsec}, the camera primarily moves straight ahead, interspersed with turns to the left or right.}
A comprehensive overview and comparison are listed in Table~\ref{tab:dataset}.
\new{As for event simulators, most of the current works~\cite{v2e,esim,voltmeter} focus on how to generate realistic events from high-frequency image sequences. However, obtaining these sequences is not easy either. While some works~\cite{video2event} propose to convert existing frame-based datasets into event data by firstly interpolating the frames densely, they are limited by the interpolation artifacts and are not suitable for low-level applications like optical flow.
In this paper, we propose BlinkSim to alleviate the problem of generating large-scale diverse scenarios with corresponding event data and optical flow labels. Both the BlinkSim and the generated data BlinkFlow are released.
}

\subsection{Event-based Optical Flow Estimation}
Extensive research has been carried out on the use of event cameras for estimating optical flow, given their identified advantages.
Previous studies have proposed adaptations of frame-based techniques (block matching~\cite{event_block_match} and Lucas-Kanade~\cite{event_LK}), 
time surface matching~\cite{event_surface},
variational optimization~\cite{event_variational},
spatio-temporal plane-fitting~\cite{plane_fitting}, contrast maximization~\cite{secrets} and so on.
These methods usually assume constant brightness.
Alternatively, another stream of research is exploring learning-based methods.
In these methods, event streams are typically converted into a grid-like representation in a pre-processing step. Then frame-based designs and frameworks such as FlowNet~\cite{flyingchair}, cost volume~\cite{pwc_net} and recurrent refinement~\cite{raft,flowformer,shi2023videoflow,shi2023flowformer++} can be applied. In total, the learning-based methods outperform the former hand-designed methods in current benchmarks~\cite{ev_flownet,dsec}.
In this paper, we propose E-FlowFormer to further improve the capacity of the learning-based method by enhancing event features through transformers.

\section{BlinkSim}

BlinkSim is a simulator designed to automate the generation of large amounts of realistic event data and high-quality optical flow ground truth.
It is mainly composed of two separate components: a configurable and photo-realistic rendering engine in a high frame rate built with the Blender\footnote{\url{https://www.blender.org}}
and a simulation suite with integrated multiple event emulators.
The two modules are decoupled and can run independently from each other. Users can simulate event data from rendered images of thousands of 3D scenes that are automatedly built from scratch using our rendering engine, or generate events from existing high-speed video data by selecting an appropriate event data simulator from our suite.
In this section, we first introduce the event generation model, then introduce the rendering engine and the event simulation module, respectively. Finally, we describe how we render the diversiform event optical flow dataset, BlinkFlow, based on the proposed simulator.

\subsection{Event Generation Model}
\new{Event cameras operate by responding to changes in the logarithmic brightness signal
in an asynchronous and independent manner for each event pixel~\cite{event_survey}.
An event is triggered when the brightness change (either increment or decrease) since the last event at that pixel reaches a threshold $\pm C$ (with $C > 0$):
}
\begin{equation}
    p_{k} (L(u_k, t_k) - L(u_k, t_k - \Delta t)) \geq C, 
\label{eq:event-generation}
\end{equation}
where $\Delta t$ is the time since the last event triggered at pixel location $u_k$ and at time $t_k$. $p_k\in \{-1,1\}$ is the polarity of the brightness change. 
To simulate the event generation process, we need access to the continuous representation of the visual signal for each pixel, which is practically unattainable. In practice, we approximate the continuous signal changes by rendering images with a high frame rate and then synthesize events according to frame-wise pixel differences.

\subsection{Rendering Engine}

The rendering engine for BlinkSim is developed with the open-source 3D creation suite Blender, based on which we can render diverse objects with complex motions under flexible scene configurations.

\noindent \textbf{3D Environment Setup}
Data-driven models benefit greatly from having access to abundant and diverse training data.
To this end, we have developed a scalable pool of rendering assets based on ShapeNet~\cite{shapenet}, which contains thousands of real-world 3D objects. The pool also contains various basic shapes like cuboids and cylinders which are randomly textured with real-world images from ADE20K dataset~\cite{ade20k}.
By default, each scene is bounded by a large textured sky box and objects are randomly selected from the pool, randomly scaled, rotated, textured, and placed on the ground plane.
The related materials can be further extended with other available resources.
For example, we can collect objects in Google Scanned Objects~\cite{google_scanned} and images in COCO~\cite{coco}.
In addition, BlinkSim provides a user-friendly interface for importing customized scenes downloaded from websites such as BlenderKit\footnote{\url{https://www.blenderkit.com/}} and Blender Store\footnote{\url{https://store.blender.org/}}.

\noindent \textbf{Trajectory Sampling}
To generate diverse motion patterns, we need to sample various trajectories, including object-moving and camera-moving trajectories.
The two trajectories are similar: for a timestamp $t$, determining a 6-degree-of-free (DoF) pose $\mathbf P(t)$ including 3-DoF rotation and 3-DoF translation.
The difference is that we put objects in the scene according to the object poses while we send camera poses to the virtual camera in Blender for rendering data.
We randomly sample N points in the free space with random rotation and fit a spline. The spline curve determines that for any timestamp $t$ we have a unique 6-DoF pose.
We also detect collisions between the objects and the camera and between multiple objects by checking the intersection of the trajectories~\cite{collision}. Once a collision is detected, we cut the trajectory of the collided objects to avoid the model breakthrough.
The next step is to sample the timestamps for rendering.
A simple strategy is sampling $t$ evenly but may fail to faithfully simulate events when the brightness signal varies more quickly than the chosen rendering frame rate.
Another choice is to adaptively increase $t$ by limiting the maximum pixel displacement or brightness change between two sampled timestamps~\cite{esim}. However, it requires generating high frame rate optical flows for pixel displacement estimation, which hinders rendering efficiency, and it is unnecessary when the motion is slow-varying.
BlinkSim supports both image sequence rendering strategies so that users can choose a suitable strategy according to their time-consuming requirements and motion settings.

\begin{figure}[!t]
\begin{center}
    \includegraphics[width=1.0\linewidth]{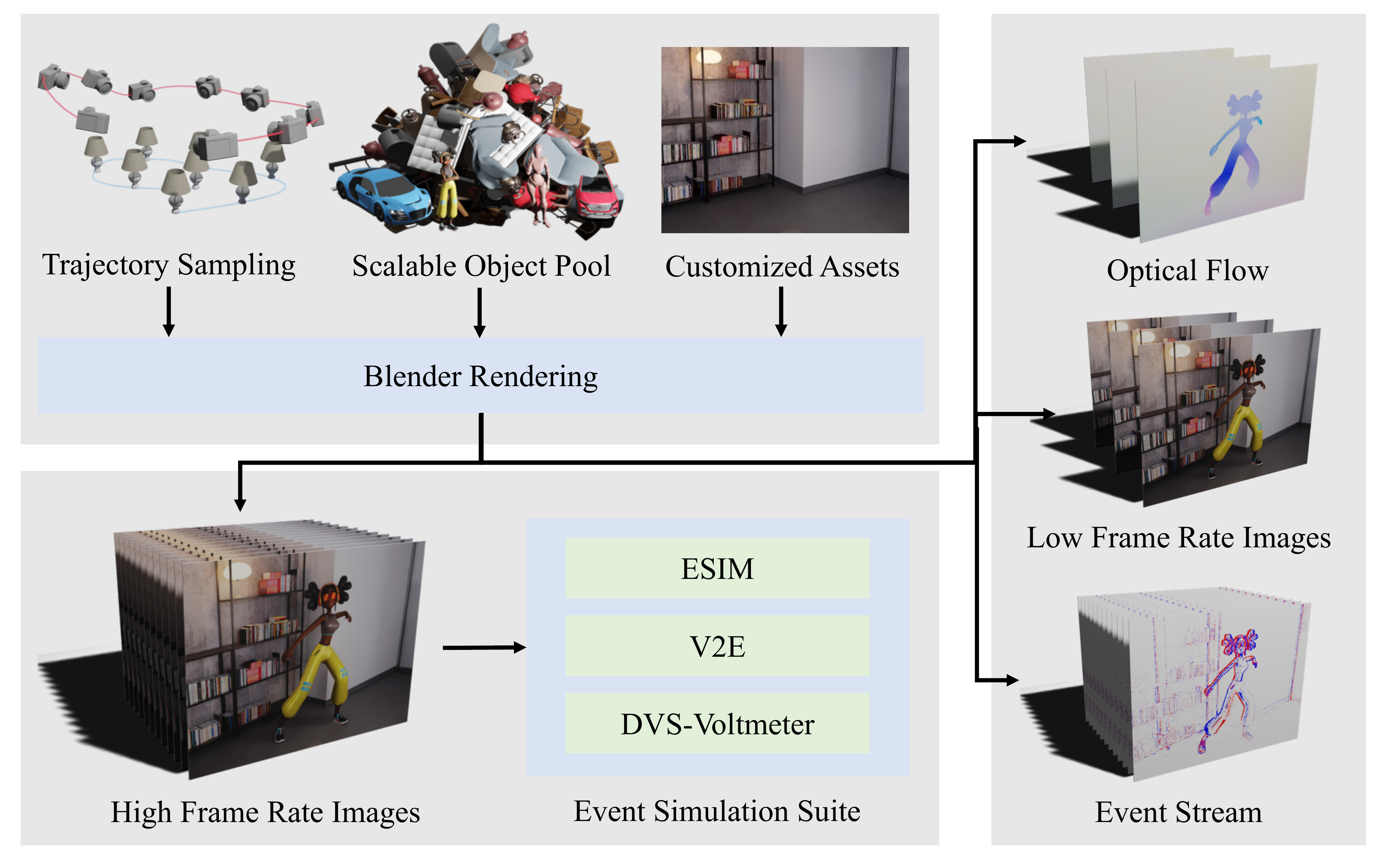}
\end{center}
\vspace{-1.0em}
\caption{Framework of BlinkSim. BlickSim consists of a configurable rendering engine and an event data simulation suite.  It allows the fast generation of large-scale data with realistic event data and related optical flow ground truth.}
\label{fig:simulator}
\vspace{-1.0em}
\end{figure}

\noindent \textbf{Data Generation}
We render the data with two passes. 
As shown in Fig.~\ref{fig:simulator}, we first render images of the densely-sampled viewpoint which are used for event data generation.
In the second pass, we sample the viewpoint sparsely and render the simulated RGB camera's view. We use BlenderProc~\cite{blenderproc} which modifies the pipeline of Blender’s internal render engine and allows us to obtain the corresponding optical flow ground truth.

\input{tables/simulator}

\subsection{Event Data Simulator}
Eq.~\ref{eq:event-generation} describes the ideal process for event generation, but in practice, the triggering threshold is not constant. Furthermore, there are complicated noise effects such as refractory periods and temporal noises at low illumination.
BlinkSim integrates three state-of-the-art (SOTA) event data simulators inside and provides flexible choices for users.
The comparison of these data simulators is listed in Table~\ref{tab:event_sim}.
ESIM~\cite{esim} is an early-stage simulator that only considers spatially varying thresholds but it runs the fastest on a CPU.
DVS-Voltmeter~\cite{voltmeter} generates the most realistic event data by considering spatiotemporally varying thresholds, modeling the event triggering timestamp as Brownian motion instead of linearly distributed. Besides, it considers the influence of temperature and parasitic photocurrent. But at the same time, the DVS-Voltmeter runs the slowest and does not have GPU acceleration.
V2E works best when simulating the scenes in low illuminations and it supports accelerating rendering with GPUs, whose speed is on par with ESIM. Users can choose suitable emulators according to the scenes and the requirement of time consumption.

\subsection{BlinkFlow Dataset}
Based on BlinkSim, we create BlinkFlow, which consists of a large-scale, diversiform training dataset and a challenging evaluation benchmark.
We split the object pool into two disjoint parts and generated 3362 and 200 scenes for training and testing, respectively.
Besides, we selected 20 sequences from the TartanAir~\cite{tartan} dataset and converted them to event data using our event simulator suite. The sequences were carefully chosen to ensure that no apparent artifacts from frame interpolation.
Finally, we downloaded some well-designed assets from Blender Store and Mixamo\footnote{\url{https://www.mixamo.com/}} and created 5 customized scenes including ``dancing women'', ``falling dice'' and so on, which are partly shown in Fig.~\ref{fig:dataset}.
\new{The resulting 200, 20, and 5 test sequences (called FlyingObjects, E-Tartan, and E-Blender, respectively) contain a total of over 4000 frames.
We choose V2E~\cite{v2e} for the event simulation when generating most of the training data and DVS-Voltmeter~\cite{voltmeter} for other training data and all the testing data.
When simulating events for each sequence, the corresponding simulation parameters are randomly selected within a range given in advance, so that the trained network do not overfit to specific parameters. For example, the contrast threshold $C$ is randomly selected in the range of [0.18, 0.25].
}
As shown in Table~\ref{tab:dataset}, BlinkFlow contains complex camera motions and dynamic objects based on our trajectory sampling, produces events with ground truth flows for occluded objects, provides individual training/test splits, covers more training scenes (3362 v.s. 18) and test scenes (225 v.s. 7), and generates more training frames (33k v.s. 8k).
Such a rich dataset can push the limits of event-based optical flow estimation and encourage researchers to explore better neural networks for event cameras.

\section{E-FlowFormer}
\begin{figure}[!t]
\begin{center}
  \includegraphics[width=1.0\linewidth]{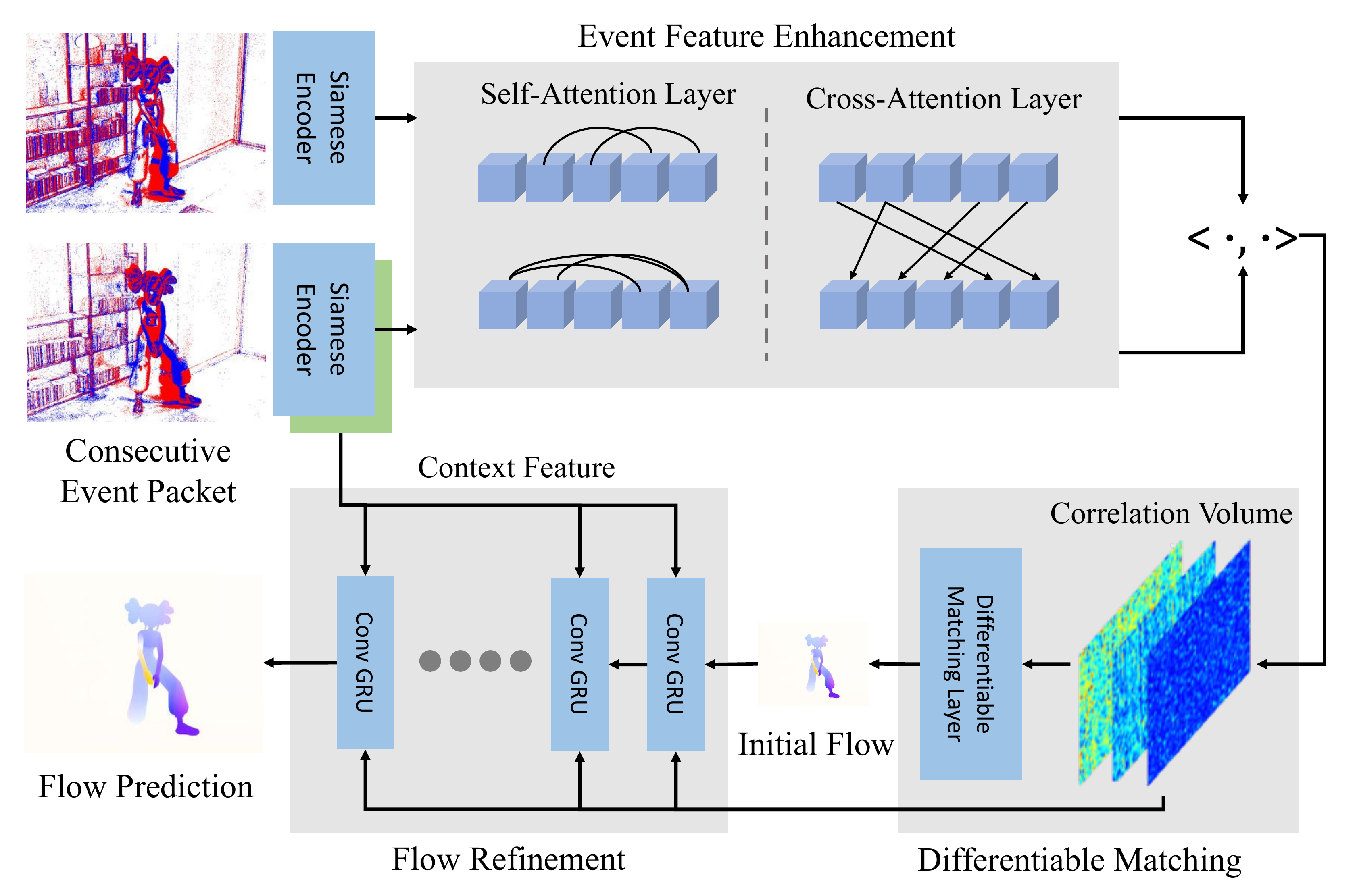}
\end{center}
\vspace{-1.0em}
\caption{Network architecture of E-FlowFormer.
The proposed feature enhancement module brings a better correlation volume through the transformer design. Besides, it provides a better initialization of optical flow for the following refinement module.
}
\vspace{-1.0em}
\label{fig:network}
\end{figure}

E-FlowFormer is built upon E-RAFT~\cite{eraft}, a state-of-the-art neural network architecture for learning event-based optical flow estimation. It encodes event features with a shallow Siamese CNN to construct a 4D correlation volume, which measures event similarity.
However, events are noisier than RGB images and the corresponding locations derived from ground truth flows do not always present the same event patterns.
Inspired by recent progress in deep learning~\cite{attention,flowformer,deltar,pats}, we propose to enhance the event feature encoding modules with transformers.
We call the enhanced neural network as E-FlowFormer.
An overview of the proposed method E-FlowFormer is presented in Fig.~\ref{fig:network}.
In this section, we first review E-RAFT and then describe our transformer-based event feature enhancement module.
\subsection{E-RAFT Brief Review}
E-RAFT consists of three stages: feature extraction, correlation volume computation, and motion refinement.
Given two consecutive packets of events, E-RAFT converts them to tensor-like event representations and then encodes local features of events $F_A, F_B \in \mathbb{R}^{H\times W\times D}$ with a siamese CNN.
With the event feature tensors $F_A$ and $F_B$, a correlation volume $C\in \mathbb{R}^{H\times W\times H\times W}$ which encodes the feature similarity for each pixel in $F_A$ with respect to all pixels in $F_B$ is computed through:
\begin{equation}
    C=\frac{F_A F_B^T}{\sqrt{D}} \in \mathbb{R}^{H\times W\times H\times W} ,
\label{eq:correlation}
\end{equation}
where $\frac{1}{\sqrt{D}}$ is a normalization factor that can avoid large values after the dot-product operation.
Then, E-RAFT iteratively refines the flow estimation by taking the current motion estimation, the correlation volume, and the context features with a ConvGRU module. The flow is usually initialized to zero in previous implementation~\cite{eraft}.
E-RAFT supervises all flow predictions using a sequence loss:
\begin{equation}
    L=\sum_{i=1}^{N} \gamma^{N-1} \|V_i-V_{gt}\|_1 ,
\label{eq:loss}
\end{equation}
where $N$ is the number of flow predictions. $\gamma$ is the weight that exponentially increases to give higher weights for later predictions. $V_i$ is the flow prediction at the $i$-th iteration and $V_{gt}$ is the ground truth flow.
\subsection{Event Feature Enhancement Through Transformers}
The key to E-RAFT lies in obtaining high-quality discriminative event features. As previously mentioned, the features $F_A$ and $F_B$ are encoded independently from a weight-sharing shallow CNN. We enhance the event feature encoding with transformers to incorporate their interdependence better.
Cooperated with positional encoding, transformers can effectively capture both the relevant spatial relationships and feature similarity between the two sets of events and thus generate refined and discriminative event features.
\new{A remaining issue is that the standard Transformer architecture introduces quadratic computational complexity.}
To avoid this issue, we adopt the spatially separable self-attention proposed in Twins~\cite{twins_transformer} instead of the vanilla version.
The enhanced event feature encoder improves the distinctiveness of event features and thus brings a better correlation volume, based on which, we can identify the correspondence through a differentiable matching layer~\cite{pats}.
The resulting correspondence will be added and supervised in the sequence loss, effectively suppressing the correlation strength far from the ground truth locations.
Besides, we use the resulting correspondence to better initialize the following refinement module.

\begin{figure*}[!t]
\begin{center}
      \resizebox{0.9\linewidth}{!}{
\includegraphics[width=1.0\linewidth]{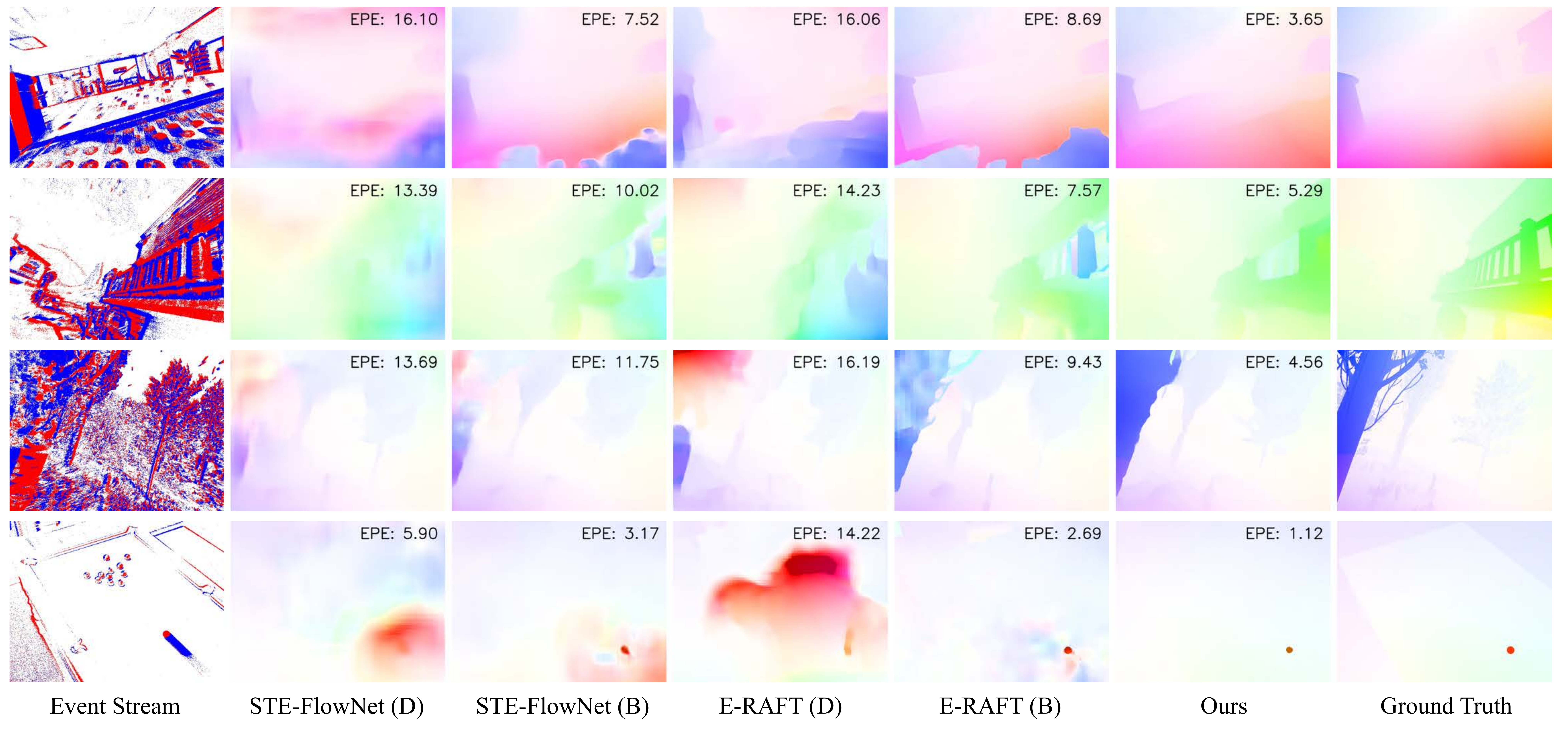}
}
\end{center}
\vspace{-1.0em}
\caption{Qualitative comparison on BlinkFlow. BlinkFlow contains challenging data that makes previous methods trained on DSEC almost fail. Models trained on BlinkFlow consistently outperform models trained on DSEC and E-FlowFormer achieves the lowest AEE among all the methods, producing accurate and smooth optical flow prediction.
}
\label{fig:comp}
\end{figure*}

\input{tables/gen}

\section{EXPERIMENT}
In this section, we first show that the models trained on BlinkFlow gain better generalization performance than those trained on DSEC and MVSEC, which reveals the superiority of training data in BlinkFlow.
Then, we demonstrate that the evaluation benchmark in BlinkFlow is more challenging and comprehensive.
Finally, we evaluate the proposed E-FlowFormer which ranks 1st on the MVSEC, DSEC and BlinkFlow benchmarks. This indicates that the transformer-based event feature enhancement module effectively learns the event features.

\noindent \textbf{Dataset}
There are two commonly used datasets for event-based optical flow training and evaluation: MVSEC~\cite{mvsec} and DSEC~\cite{dsec}.
MVSEC contains two scenarios, i.e., sequences recorded in the indoor room and outdoor sequences recorded while driving on public roads. The ground truth of optical flow is approximated by multiplying the motion field by the time interval (denoted as ``dt'') where the motion field is computed according to the camera movement and scene depth.
DSEC~\cite{dsec} contains data in a higher resolution and improves the quality of the ground truth optical flow by removing the occlusions and moving objects.
Both of MVSEC and DSEC collect data in rigid and static scenes and compute optical flows via post-processing, resulting in highly biased data and inaccurate ground truth.
In contrast, our BlinkFlow dataset covers various scenes and objects and directly obtains the ground truth flows from the renderer.
We denote DSEC, MVSEC, and BlinkFlow datasets by ``D'', ``M'', and ``B''.

\noindent \textbf{Experiment Setup.}
\new{
We measure the accuracy of the optical flow using several metrics, namely the Average Endpoint Error (AEE), outlier rate, Angular Error (AE), and N-pixel errors (N-PE).
The AEE measures the Euclidean distance between the predicted flow and the ground truth.
The outlier rate represents the percentage of points where the endpoint error exceeds 3 pixels and 5\% of the magnitude.
The AE calculates the angle between normalized optical flow predictions and the normalized ground truth.
N-PE denotes the percentage of flow errors greater than N pixels in magnitude.
}

\input{tables/blinkflow}
\input{tables/overfit_dataset}

\noindent \textbf{Training Data in BlinkFlow Improves Generalization.}
We set up two experiments with three network architectures, i.e., STE-FlowNet~\cite{ste_flownet}, E-RAFT~\cite{eraft}, and our E-FlowFormer, to present that models trained on BlinkFlow (B) gain the best generalization performance.
We train models on MVSEC (M) and BlinkFlow (B) and evaluate the models on DSEC.
As shown in Table~\ref{tab:gen}, the models trained on BlinkFlow achieve at least 84\% error reduction.
Then, we further train the models on DSEC and BlinkFlow and evaluate them on MVSEC (Table~\ref{tab:gen}), which also reveals significant performance improvement for E-RAFT and our E-FlowFormer, up to 34\% AEE reduction on MVSEC (indoor).
BlinkFlow does not outperform DSEC much when evaluated on the outdoor sequence of MVSEC because DSEC and MVSEC (outdoor) share a similar scene pattern, i.e., both are outdoor driving scenarios.
Totally speaking,
both experiments present the extraordinary superiority of our BlinkFlow.
Moreover, our proposed E-FlowFormer makes the best use of BlinkFlow and outperforms the other two methods all-sided.

\noindent \textbf{Test Data in BlinkFlow Is More Challenging.}
All of DSEC, MVSEC, and our proposed BlinkFlow provide both training data and test data.
We can train and test models with data belonging to the same dataset to present the easiness of datasets.
We select E-RAFT and STE-FlowNet and use outlier metrics in this experiment.
As shown in Table~\ref{tab:overfit_dataset},
E-RAFT and STE-FlowNet obtain overall less than 4 percent in terms of outliers, which indicates that DSEC and MVSEC are so simple that they can be easily fitted.
However, BlinkFlow is still challenging for them, which reveals that BlinkFlow is a more comprehensive evaluation benchmark to push the limits of event-based optical flow estimation.

\input{tables/mvsec}

\noindent \textbf{E-FlowFormer Ranks 1st on MVSEC.}
In Table~\ref{tab:mvsec}, we compare with recent SOTA methods on the MVSEC dataset.
E-FlowFormer, only trained on the proposed BlinkFlow dataset achieves comparable performances with the SOTA methods leveraging the MVSEC training data. Especially, E-FlowFormer outperforms other methods by a significant margin in almost all the indoor sequences, because the training set of MVSEC is only the outdoor sequences.
It is easy to overfit the training data and thus brings severe performance degradation on indoor sequences. On the contrary, BlinkFlow covers complex scene structures so our method can generalize well on both indoor and outdoor sequences.
We also fine-tune E-FlowFormer on the data combined from BlinkFlow and MVSEC (denoted as ``B+M'').
Fine-tuning further improves the performance of E-FlowFormer, especially on the sequence of outdoor day1.
Finally, E-FlowFormer outperforms all the published methods in almost all the sequences, up to 28\% in terms of AEE metric and 91\% in terms of outlier metric.

\input{tables/dsec}

\noindent \textbf{E-FlowFormer Ranks 1st on DSEC.}
We show the comparisons on the DSEC dataset in Table~\ref{tab:dsec}.
Only trained on the proposed BlinkFlow dataset, E-FlowFormer surpasses most published methods except for E-RAFT.
After fine-tuning on the data combined from BlinkFlow and DSEC (denoted as ``B+D''), E-FlowFormer outperforms all other methods in terms of all metrics, and up to 14\% in terms of 2PE metric.
Notably, the margin between our approach and SOTA methods is smaller on DSEC than on MVSEC, because DSEC provides much more training data than MVSEC. Meanwhile, both the training and test data of DSEC are collected in outdoor driving scenarios, making it easy to overfit the testing data.

\noindent \textbf{E-FlowFormer Ranks 1st on BlinkFlow.}
We train several state-of-the-art event-based optical flow estimation methods with the training data in BlinkFlow and test them on the test split in BlinkFlow, as shown in Table~\ref{tab:custom}.
E-FlowFormer outperforms previous methods all-sided.
E-Blender, which contains flexible objects, is the most challenging subset while our E-FlowFormer obtains the most significant AE improvement in this subset.
It denotes that E-FlowFormer best exploits the sufficient training data provided in BlinkFlow.
More qualitative results can be found in Fig.~\ref{fig:comp}.

\section{CONCLUSION}

In this paper, we have proposed a novel simulator, BlinkSim, and a large-scale and diversiform dataset, BlinkFlow.
Thanks to the well-integrated BlinkSim, BlinkFlow is two orders of magnitude ahead of other available datasets in diversity, so it provides a much better training and test set.
We believe that BlinkFlow will push the limits of event-based optical flow estimation and BlinkSim can also benefit other event-based tasks.
Moreover, we propose a novel transformer-based event flow estimation network, E-FlowFormer, which effectively learns from the training set of BlinkFlow and ranks 1st on mainstream benchmarks DSEC and MVSEC and the proposed BlinkFlow benchmark.

\bibliographystyle{IEEEtranS}
\bibliography{IEEEabrv}

\end{document}

%% file: tables/dataset.tex
\begin{table*}[!t]
\centering

\caption{Comparison of available datasets.}
\begin{tabular}{lccccccc}
\toprule
Dataset & Motion Pattern & Dynamic Object & Occlusion & Training Frames & Training Scenes & Test Scenes & Resolution \\ \hline
DVSFLOW16~\cite{DVSFLOW16} & Rotation & \xx & \xx & - & - & 5 & 180 $\times$ 240 \\
DVSMOTION20~\cite{event_surface} & Rotation & \xx & \xx & - & - & 4 & 260 $\times$ 346 \\
MVSEC~\cite{mvsec} & Drone & \xx & \xx & 3k & 1 & 4 & 260 $\times$ 346 \\
DSEC~\cite{dsec} & Car & \xx & \xx & 8k & 18 & 7 & 640 $\times$ 480 \\
BlinkFlow (Ours) & Random & \cc & \cc & 33k & 3362 & 225 & 640 $\times$ 480 \\ \bottomrule
\end{tabular}
% \vspace{-15px}
% \vspace{-3.0em}
\label{tab:dataset}
\end{table*}

%% file: tables/simulator.tex
\begin{table*}[!t]
\centering
\tabcolsep 3pt
\footnotesize
% \\xxrmalsize

\caption{Comparison of three event data simulator.}
\begin{tabular}{ccccccc}
\toprule
 & Threshold & \makecell{Timestamp\\Modeling} & \makecell{Temperature\\Noise} & \multicolumn{1}{l}{\makecell{Low\\Illumination}} & \multicolumn{1}{l}{\makecell{Refractory\\Periods}} & \makecell{Speed (10k ev./s)\\ CPU / GPU} \\ \hline
ESIM~\cite{esim} & Spatial & Linear & \xx & \xx & \xx & 170 / -\\
V2E~\cite{v2e} & Spatial & Linear & \xx & \cc & \cc & 26 / 141 \\
\makecell{DVS-Voltmeter\cite{voltmeter}} & \multicolumn{1}{l}{Spaitio-temperoal} & Brownian & \cc & \xx & \xx & 29 / -\\ \bottomrule
\end{tabular}

\label{tab:event_sim}
% \vspace{-15px}
\vspace{-1.0em}
\end{table*}

%% file: tables/gen.tex
\begin{table*}[!t]
\centering
\tabcolsep 7pt
\footnotesize
% \normalsize

\caption{Generalization performance comparison on DSEC and MVSEC dataset.}
\begin{tabular}{lcccccccc}
\toprule
  &  & \multicolumn{2}{c}{DSEC (train)} &  & \multicolumn{2}{c}{MVSEC (indoor)} & \multicolumn{2}{c}{MVSEC (outdoor)} \\ \cline{3-4} \cline{6-9} 
\multirow{-2}{*}{Methods} & \multirow{-2}{*}{Training Data} & AEE & Outlier & \multirow{-2}{*}{Training Data} & AEE & Outlier & AEE & Outlier \\ \hline
 & M & 9.91 & 77.59 & D & 2.17 & 19.77 & 1.34 & 8.52 \\
 & B & 1.45 & 9.19 & B & 1.72 & 12.21 & 1.37 & 8.59 \\
\multirow{-3}{*}{\makecell{STE-FlowNet\\\cite{ste_flownet}}} & \cellcolor{impro}Impro. & \cellcolor{impro}+85\% & \cellcolor{impro}+88\% & \cellcolor{impro}Impro. & \cellcolor{impro}+22\% & \cellcolor{impro}+38\% & \cellcolor{impro}-2\% & \cellcolor{impro}-1\% \\ \hline
 & M & 11.24 & 84.89 & D & 2.29 & 18.12 & 1.47 & 8.98 \\
 & B & 1.34 & 8.12 & B & 1.85 & 12.03 & 1.35 & 8.56 \\
\multirow{-3}{*}{E-RAFT~\cite{eraft}} & \cellcolor{impro}Impro. & \cellcolor{impro}+88\% & \cellcolor{impro}+90\% & \cellcolor{impro}Impro. & \cellcolor{impro}+19\% & \cellcolor{impro}+34\% & \cellcolor{impro}+8\% & \cellcolor{impro}+5\% \\ \hline
 & M & 7.92 & 65.85 & D & 2.44 & 22.35 & 1.45 & 9.21 \\
 & B & 1.25 & 6.73 & B & 1.62 & 11.18 & 1.29 & 8.08 \\
\multirow{-3}{*}{\makecell{E-FlowFormer\\(Ours)}} & \cellcolor{impro}Impro. & \cellcolor{impro}+84\% & \cellcolor{impro}+90\% & \cellcolor{impro}Impro. & \cellcolor{impro}+34\% & \cellcolor{impro}+50\% & \cellcolor{impro}+11\% & \cellcolor{impro}+12\% \\ \bottomrule
\end{tabular}

\label{tab:gen}
% \vspace{-10px}
\vspace{-1.0em}
\end{table*}

%% file: tables/blinkflow.tex
\begin{table*}[!t]
\centering
\tabcolsep 7pt
% \scriptsize
% \normalsize
\footnotesize

\caption{Comparison on BlinkFlow Dataset.}
    % \resizebox{0.8\linewidth}{!}{
\begin{tabular}{lccccccccc}
\toprule
\multirow{2}{*}{Methods} & \multicolumn{3}{c}{FlyingObjects} & \multicolumn{3}{c}{E-Tartan} & \multicolumn{3}{c}{E-Blender} \\ \cline{2-10} 
 & AEE & Outlier & AE & AEE & Outlier & AE & AEE & Outlier & AE \\ \hline
Spike-FlowNet~\cite{spike_flownet} & 5.39 & 32.16 & 11.67 & 4.13 & 31.31 & 15.26 & 3.82 & 34.72 & 28.03 \\
EV-FlowNet~\cite{ev_flownet} & 4.02 & 16.64 & 7.14 & 3.41 & 22.17 & 11.31 & 3.24 & 26.95 & 22.46 \\
E-RAFT~\cite{eraft} & 3.19 & 9.27 & 5.48 & 2.83 & 16.43 & 10.13 & 2.66 & 19.90 & 18.54 \\
STE-FlowNet~\cite{ste_flownet} & 3.22 & 10.44 & 5.97 & 2.79 & 16.80 & 9.83 & 2.59 & 18.87 & 17.43 \\
E-FlowFormer (Ours) & \textbf{2.94} & \textbf{9.03} & \textbf{4.92} & \textbf{2.48} & \textbf{15.98} & \textbf{8.42} & \textbf{2.38} & \textbf{17.56} & \textbf{14.78} \\ \bottomrule
\end{tabular}
% }
% \vspace{-10px}
\vspace{-1.0em}
\label{tab:custom}
\end{table*}

%% file: tables/overfit_dataset.tex
\begin{table}[!t]
\centering
\footnotesize
% \normalsize
\caption{Dataset Comparison by Training and Testing Models on the same dataset in Terms of outlier.}
\begin{tabular}{lccc}
\toprule
Methods & DSEC & MVSEC & BlinkFlow \\ \hline
E-RAFT~\cite{eraft} & 2.68 & 0.92 & 15.20 \\
STE-FlowNet~\cite{ste_flownet} & 4.06 & 0.50 & 15.37 \\ \bottomrule
\end{tabular}
\label{tab:overfit_dataset}
% \vspace{-2.0em}
\vspace{-0.5em}
\end{table}

%% file: tables/mvsec.tex
\begin{table}[!t]
\centering
\tabcolsep 5pt
\footnotesize
% \normalsize
\caption{Comparison on MVSEC dataset.}
\begin{tabular}{lccccc}
\toprule
\multirow{2}{*}{dt=1 frame} & \multirow{2}{*}{\makecell{Training\\Data}} & \multicolumn{2}{c}{indoor} & \multicolumn{2}{c}{outdoor} \\ \cline{3-6} 
 &  & AEE & Outlier & AEE & Outlier \\ \hline
EV-FlowNet~\cite{ev_flownet} & M & 1.43 & 9.73 & 0.49 & 0.20 \\
Spike-FlowNet~\cite{spike_flownet} & M & 1.08 & 3.90 & 0.47 & \textbf{0.00} \\
E-RAFT~\cite{eraft} & M & 0.97 & 2.76 & 0.65 & 2.19 \\
Zhu et al.~\cite{unsupervise_zhu} & M & 0.82 & 2.33 & \underline{0.32} & \textbf{0.00} \\
STE-FlowNet~\cite{ste_flownet} & M & \underline{0.69} & 1.00 & 0.42 & \textbf{0.00} \\
E-FlowFormer (Ours) & B & 0.71 & \underline{0.94} & 0.67 & 1.68 \\
E-FlowFormer (Ours) & B+M & \textbf{0.51} & \textbf{0.09} & \textbf{0.29} & 0.05 \\ \hline
\multirow{2}{*}{dt=4 frame} & \multirow{2}{*}{\makecell{Training\\Data}} & \multicolumn{2}{c}{indoor} & \multicolumn{2}{c}{outdoor} \\ \cline{3-6} 
 &  & AEE & Outlier & AEE & Outlier \\ \hline
EV-FlowNet~\cite{ev_flownet} & M & 3.25 & 36.57 & 1.23 & 7.30 \\
Spike-FlowNet~\cite{spike_flownet} & M & 3.08 & 33.43 & 1.09 & 5.60 \\
E-RAFT~\cite{eraft} & M & 2.46 & 23.15 & 1.43 & 9.17 \\
Zhu et al.~\cite{unsupervise_zhu} & M & 3.07 & 39.60 & 1.30 & 9.70 \\
STE-FlowNet~\cite{ste_flownet} & M & 2.17 & 20.97 & \underline{0.99} & \underline{3.90} \\
E-FlowFormer (Ours) & B & \underline{1.62} & \underline{11.18} & 1.29 & 8.08 \\
E-FlowFormer (Ours) & B+M & \textbf{1.56} & \textbf{10.25} & \textbf{0.83} & \textbf{3.44} \\ \bottomrule
\end{tabular}
% \vspace{-15px}
\vspace{-0.5em}
\label{tab:mvsec}
\end{table}

%% file: tables/dsec.tex
\begin{table}[!t]
\centering
\tabcolsep 4pt
\footnotesize
% \normalsize

\caption{Comparison on DSEC dataset.}
\begin{tabular}{lcccccc}
\toprule
Methods & \makecell{Training\\Data} & 1PE & 2PE & 3PE & AEE & AE \\ \hline
MultiCM~\cite{secrets} & D & 76.57 & 48.48 & 30.86 & 3.47 & 13.98 \\
OF-EV-SNN~\cite{of_snn} & D & 53.67 & 20.24 & 10.31 & 1.71 & 6.34 \\
E-RAFT~\cite{eraft} & D & 12.74 & 4.74 & 2.68 & 0.79 & 2.85 \\
E-FlowFormer(Ours) & B & 36.39 & 13.02 & 6.05 & 1.33 & 4.65 \\
E-FlowFormer(Ours) & B+D & \textbf{11.23} & \textbf{4.10} & \textbf{2.45} & \textbf{0.76} & \textbf{2.68} \\ \bottomrule
\end{tabular}

\label{tab:dsec}
\vspace{-1.0em}
\end{table}